\documentclass[11pt]{article}

\usepackage[preprint]{acl}

\usepackage{times}
\usepackage{latexsym}

\usepackage[T1]{fontenc}

\usepackage[utf8]{inputenc}

\usepackage{microtype}

\usepackage{inconsolata}

\usepackage{graphicx}

\usepackage{amsmath}
\usepackage{amssymb}
\usepackage{wrapfig}
\usepackage{multirow}
\usepackage{booktabs}
\usepackage{enumitem}
\usepackage{algorithm}
\usepackage{algpseudocode}
\usepackage{amsmath}
\usepackage{colortbl} 
\usepackage[most]{tcolorbox} 
\usepackage{tcolorbox}
\tcbuselibrary{listingsutf8}
\usepackage[table]{xcolor}
\usepackage{tcolorbox}
\usepackage{multicol}
\usepackage{xspace}
\usepackage{array}
\usepackage[dvipsnames, table]{xcolor}
\tcbuselibrary{breakable, skins}
\newcommand{\ours}{\textbf{SLKD}\xspace}

\title{Logical Structure as Knowledge: Enhancing LLM Reasoning via Structured Logical Knowledge Density Estimation}

\author{
 \textbf{Zhen Bi\textsuperscript{1,2,3*}},
 \textbf{Zhenlin Hu\textsuperscript{1,2}$\thanks{Equal Contribution.}$},
 \textbf{Xueshu Chen\textsuperscript{1,2}},
 \textbf{Mingyang Chen\textsuperscript{4}},
 \textbf{Cheng Deng\textsuperscript{3,5}},
 \\
 \textbf{Yida Xue\textsuperscript{6}},
 \textbf{Zhen Wang\textsuperscript{1,2}},
 \textbf{Qing Shen \textsuperscript{1,2}},
 \textbf{Ningyu Zhang\textsuperscript{6}},
 \textbf{Jungang Lou\textsuperscript{1,2$\thanks{Corresponding Author.}$}},
\\
 \textsuperscript{1}Huzhou University,
 \textsuperscript{2}Zhejiang Key Laboratory of Intelligent Education Technology and Application,
\\
 \textsuperscript{3}Banbu AI Foundation,
 \textsuperscript{4}Baichuan Inc,
 \textsuperscript{5}University of Edinburgh,
 \textsuperscript{6}Zhejiang University
\\
   \href{bizhen_zju@zju.edu.cn} {bizhen\_zju@zju.edu.cn}
}

\begin{document}
\maketitle
\begin{abstract}

The reasoning capabilities of Large Language Models (LLMs) are increasingly attributed to training data quality rather than mere parameter scaling. However, existing data-centric paradigms often equate quality with factuality or diversity and ignore the internal logical complexity of training samples. 
In this work, we propose that natural language harbors \textit{Structured Logical Knowledge } manifested through entailment relationships and logical topologies.
To quantify this, we introduce Structured Logical Knowledge Density (\ours), a novel metric that measures logical information content by decomposing natural language into executable predicates and logical primitives. Our analysis reveals a significant logical disparity in current datasets where sparse logical signals predominate. 
Consequently, we propose a \textit{density-aware re-cognizing optimization} strategy that prioritizes high-density logical samples to enhance with the LLM's reasoning ability. 
Extensive experiments demonstrate that our approach enhances reasoning performance and generalization without increasing total data volume. These results, further validated within a reinforcement learning framework, suggest that elevating logical density is more critical than expanding data scale for realizing the full cognitive potential of LLMs.
The released code is available in the Appendix~\ref{appendix:exp_details}.


\end{abstract}

\section{Introduction}

Recent advances in Large Language Models (LLMs) are attributed not merely to the scaling of parameters but significantly to the enhancement of training data quality.
Within the data-centric paradigm \citep{brown2020language,DBLP:journals/jmlr/RaffelSRLNMZLL20,hoffmann2022trainingcomputeoptimallargelanguage,DBLP:conf/nips/ZhouLX0SMMEYYZG23}, a consensus has emerged: high-quality data serves as the cornerstone of LLM reasoning capabilities. 

\begin{figure}[ht]
    \centering
    \includegraphics[width=0.95\linewidth]{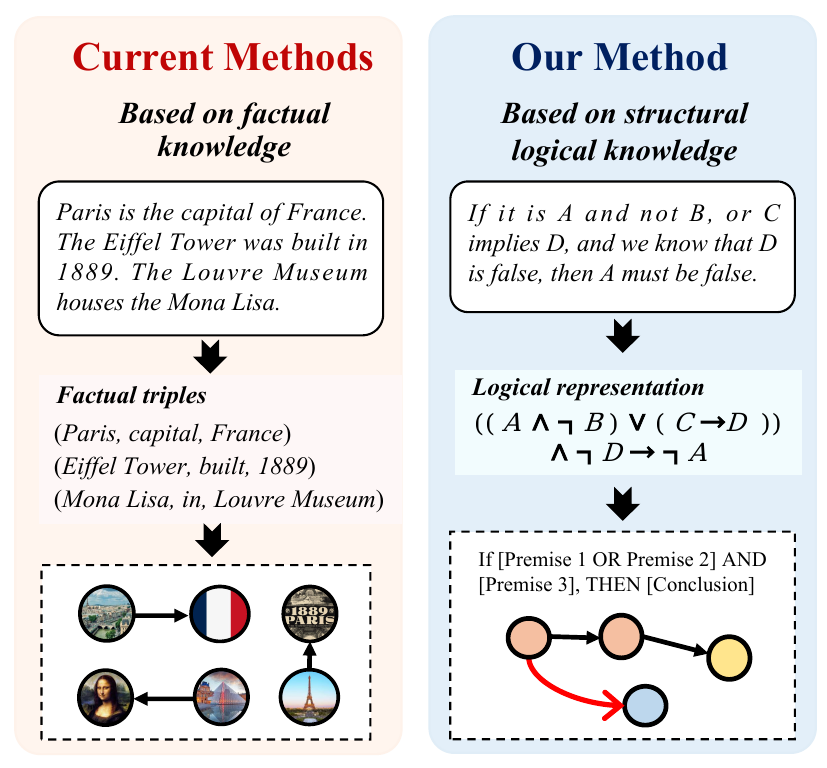}
    \caption{
    Comparison between the current methods and our method.
    The text \textbf{on the left} contains entities and relationships, but it is logically quite straightforward.
    The logical structure of the text \textbf{on the right} is very dense, but it contains many meaningless symbols.
    }
    \label{fig:motivation_fig}
    \vspace{-7mm}
\end{figure}

However, the precise definition of data quality remains elusive. Prevailing research predominantly focuses on dimensions such as factuality and diversity\citep{hua2025attentioninfluenceadoptingattentionhead,wang2025datawhispererefficientdata}. 
From this perspective, knowledge is often equated with entities and facts stored within the text, and we argue that this definition is incomplete.
Beyond declarative facts, natural language implicitly harbors another crucial form of knowledge:  \textbf{Structured Logical Knowledge}.
This knowledge manifests not as isolated entities, but as entailment relationships between predicates, conditional constraints, and logical topological structures.


Although there are currently some studies on the complexity of training data for LLMs \citep{DBLP:conf/nips/PrystawskiLG23,DBLP:conf/aaai/Bi0JDZC24}, we have long lacked an effective methodology to quantify the precise knowledge density contained within data. 
Existing analytical approaches typically assess knowledge richness by analyzing factual triples implicitly contained in the text or the underlying syntactic trees. 
However, they struggle to model the distinctions in knowledge density inherent to texts featuring complex deductive reasoning (shown in Figure \ref{fig:motivation_fig}). 
This absence of appropriate metrics often leads to an indiscriminate expansion of data volume, which may overlook the critical role of logical structure density as a decisive variable in the emergence of model reasoning capabilities.


To bridge this theoretical gap, we propose regarding logical structure as a measurable form of knowledge. 
We introduce a novel metric: \textit{Structured Logical Knowledge Density} (\ours).
Distinct from traditional metrics for text knowledge complexity, \ours does not focus on the factual knowledge contained in the data, but instead delves into the logical core.
By decomposing natural language into executable predicates and logical primitives, and calculating the topological complexity of their combinatorial reasoning processes, it objectively measures the logical information content inherent in a sample.

Leveraging \ours, we identify a significant logical disparity in existing training data distributions—a vast amount of data provides only sparse logical knowledge, while the importance of \textbf{high-SLKD} samples is overlooked. 
To address this, we propose a \textit{density-aware re-cognizing optimization strategy}. 
Instead of merely pursuing data scale, this strategy reshapes the training process according to \ours: it first establishes a foundational logical schema via extensive density-based data, and subsequently concentrates computational resources on high-density samples, thereby enabling the model to efficiently acquire complex logical entailment patterns.
The main contributions  are as follows:

\begin{itemize}[leftmargin=0.3cm]
    \item  We propose the perspective of  
    \textit{Logical Structure as Knowledge}, highlighting the oversight of logical knowledge density in current data estimation paradigms.
    \item  We introduce \ours, an objective metric for quantifying the complexity of internal logical entailment, capable of precisely identifying "high logical value regions" within data.
    \item  We design an optimization strategy based on \ours that significantly enhances LLM performance on logical reasoning tasks without increasing the total data volume. This confirms that the key to improving model reasoning lies in elevating the logical knowledge density of training data, rather than merely expanding its scale.
\end{itemize}

\section{Related Work}
\subsection{Data-Centric Reasoning for LLMs}
Recent work increasingly emphasizes that, beyond model architecture and algorithm design, the structure, provenance, and difficulty of training data critically shape LLM reasoning performance \citep{long_cot_survey,survey-efficient-reasoning-large,DBLP:journals/corr/abs-2503-16419,ruis2025pretraining}. This data-centric shift prioritizes data composition and quality over sheer volume to enhance reasoning ability \citep{When-More-is-Less,DBLP:conf/acl/JinYSZHMZD24,wang2025scale}.
Many studies highlight that higher-quality, logically structured examples—rather than simply more data—yield better generalization and reasoning performance \citep{DBLP:journals/corr/abs-2502-19363,yu2025rethinkinggenerationhighqualitycot,li2025distilling,wettig2024qurating,zhao-etal-2024-decoratelm,yu2024mates}. In particular, aligning sample difficulty with model capability is identified as key to effective training \citep{DBLP:journals/corr/abs-2502-09650}. Instruction tuning research supports this view, showing that both prompt quality and exposure timing influence reasoning emergence \citep{qingsong2025raisereinforencedadaptiveinstruction,kim2024data}. Meanwhile, \citet{kandpal2025positionexpensivellmtraining} highlight the often-overlooked human labor cost in curating such training data.

\begin{figure*}[hbtp]
    \centering
    \includegraphics[width=1.0\textwidth]{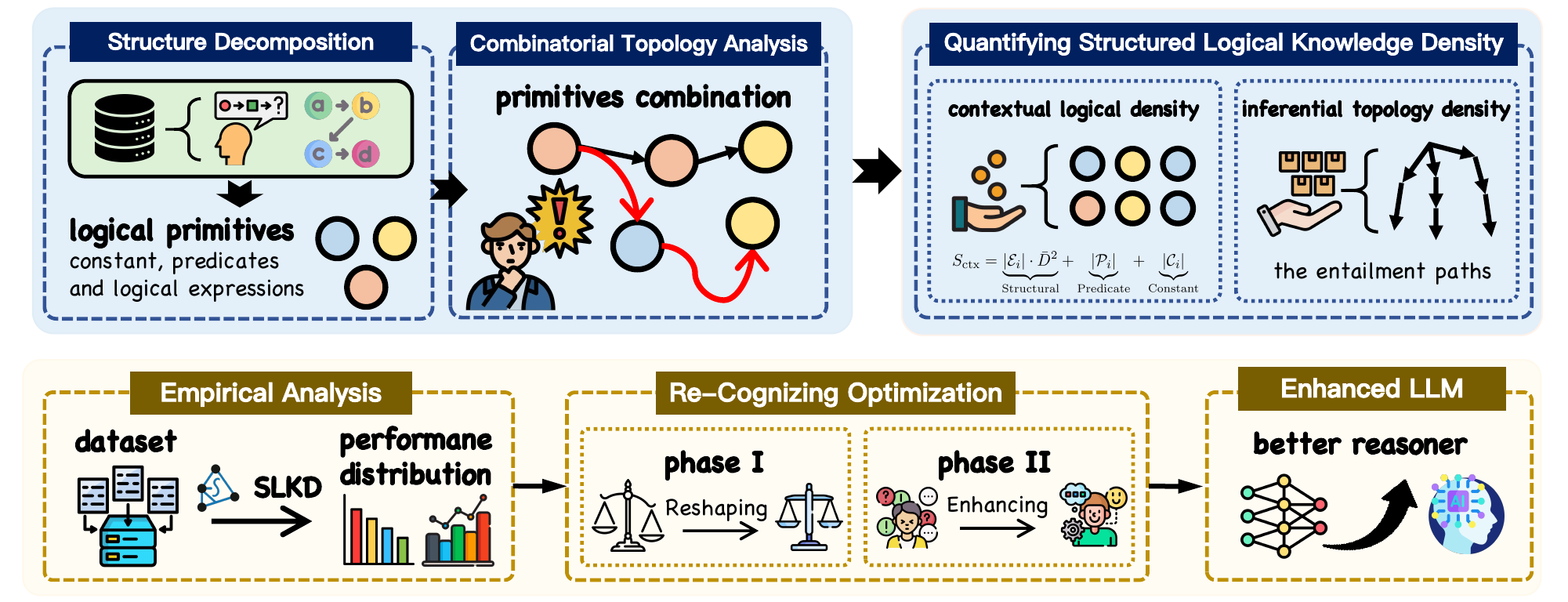}
    \caption{
    The overall framework.
    \textbf{Top: Logical structure as knowledge}. 
    We first extract logical elements from each example and perform combinatorial reasoning to derive  \textit{Structured Logical Knowledge Density} (\ours) score. \textbf{Bottom: SLKD-based optimization method}.
    We then analyze the performance distribution across \ours levels on multiple datasets. Based on this, we propose the \textit{re-cognizing optimization} strategy: the first stage reshapes the model’s recognition of reasoning patterns, and the second enhances its logical reasoning capability, thereby improving overall performance.}
    \label{fig:framework}
    \vspace{-5mm}
\end{figure*}

\subsection{Reasoning Emergence for LLMs}
Complementing these data-centric strategies, other studies explore how reasoning emerges from the interaction between cognition and structured input.
For instance, stepwise reasoning is viewed as an emergent property of sequential data exposure rather than a fixed architectural feature 
\citep{DBLP:conf/nips/PrystawskiLG23}. Furthermore, only data within a suitable complexity range appears to effectively stimulate reasoning behavior \citep{DBLP:conf/aaai/Bi0JDZC24}, suggesting that model performance is bounded by cognitive processing capacity. Empirical evidence further shows that 
carefully curated datasets often outperform larger but noisier corpora in supporting reasoning skills \citep{DBLP:journals/corr/abs-2502-12853,DBLP:journals/corr/abs-2502-03387,morishita2024fld,wang2025ultrafinewebefficientdatafiltering,yang2025iconincontextcontributionautomatic,hua2025attentioninfluenceadoptingattentionhead}, reinforcing the value of reasoning supervision that is both selective and structurally rich.
Prior work underscores data’s role in LLM reasoning but often lacks precise difficulty metrics and clear goals. We introduce \ours, a unified score for reasoning potential, and \textit{re-cognizing optimization}, which emphasizes high \ours examples while preserving diversity to enhance LLM's logical reasoning performance.

\section{Logical Structure as Knowledge}\label{sec:Methodology}

In this section, we introduce \textit{Structured Logical Knowledge Density} (\ours).
The calculation of \ours involves transforming unstructured natural language into structured predicate logic representations and then evaluating the topological complexity of the resulting reasoning chains.




\subsection{Structure Decomposition}
The first step in quantifying \ours is to decouple the logical form from the natural language surface. We treat the reasoning process as a manipulation of executable predicates. Given a data sample $Q$, we employ a distillation function $f$ (implemented via advanced LLMs) to decompose the text into three fundamental \textbf{logical primitives}:

\begin{equation}
    f(Q) \Rightarrow \{P, C, E\}
\end{equation}

\begin{itemize}[
    label=\raisebox{0.25ex}{\scalebox{0.6}{$\bullet$}},
    leftmargin=1em,   
    itemindent=0pt,   
    noitemsep,        
    topsep=0pt        
]
    \item \textbf{Constants}: $C$ is the set of entities or values involved in the reasoning context (e.g., specific objects, names).
    
    \item \textbf{Predicates}: $P$ is the set of relations or properties binding the constants (e.g., $\texttt{IsParentOf}(x, y)$, $\texttt{Greater}(a, b)$).
    
    \item \textbf{Logical Expressions}: $E$ is the set of explicit logical constraints or rules from the context, formalized as first-order logic expressions.
\end{itemize}

This decomposition converts the fuzzy information of natural language into a rigorous \textbf{predicate knowledge base}, serving as the substrate for subsequent density analysis.

\subsection{Combinatorial Topology Analysis}

The density of logical knowledge arises from \textbf{how these primitives combine to form valid entailment paths}, we define the combinatorial topology analysis process to reconstruct the logical trajectory required to derive the answer from the primitives.
Using a function $F$, we generate the reasoning topology, which consists of the precondition structure $\bar{E}$ and a sequence of derivation steps $S$:

\begin{equation}
    \text{Topology} = F \left( \text{Primitives}, A \right) \Rightarrow \{\bar{E}, \text{S}\}
\end{equation}

where $A$ is answer set, $\bar{E}$ is the precondition logic structure and $S$ is the single reasoning chain required for the reasoning trajectory. Detailedly, each reasoning node  $s_k \in S$ contains:

\begin{equation}
    s_k = \left( \#\text{Operations}_k, D^{\text{nest}}_k, \text{Expression}_k \right)
\end{equation}

where $\#\text{Operations}_k$ counts the logical operators (\texttt{AND}/\texttt{OR}/\texttt{NOT}), $D^{\text{nest}}_k$ represents the nesting depth of the logical expression, and \text{Expression}\_k is its formal representation. 

The nesting depth is a critical parameter, as it reflects the hierarchical complexity of the entailment relationship.

\begin{figure*}[hbt!]
    \centering
    \includegraphics[width=1.0\textwidth]{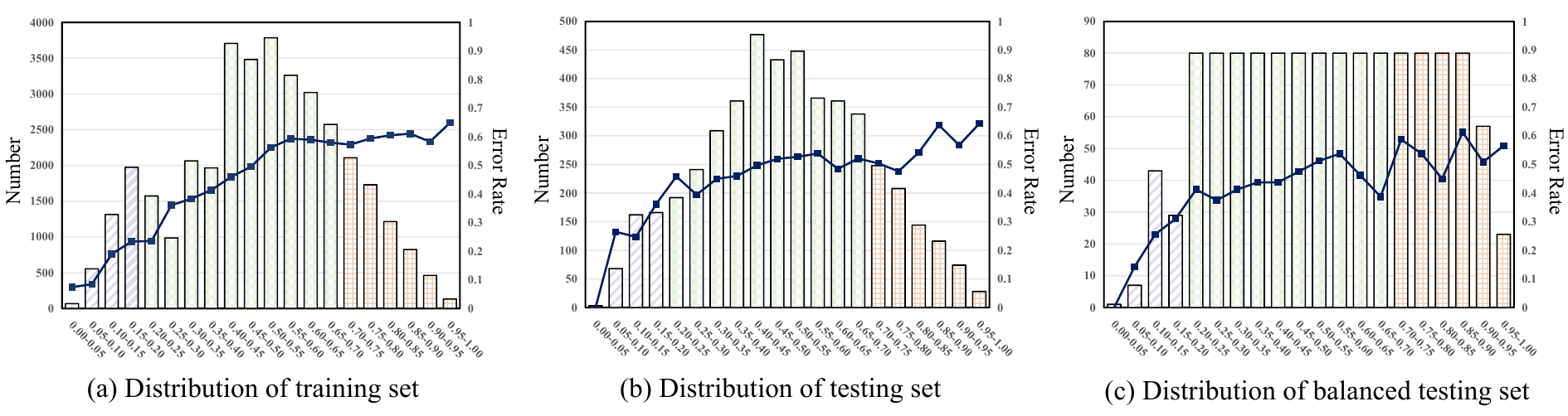}
    \caption{
    Effectiveness verification for \ours. Sample counts (bars, left axis) and model error rates (lines, right axis) are shown across \ours score bins.
    All three panels share the same layout: the x-axis divides the score range into uniform intervals, the bar height indicates the number of examples per interval, and the overlaid line traces the error rate. 
    (a) Training-set distribution and error. 
    (b) Original test-set distribution.
    (c) Balanced test-set distribution.}
    \label{fig:Data_Distribution_2}
    \vspace{-6mm}
\end{figure*}

\subsection{Quantifying Structured Logical Knowledge Density (\ours)}


We quantify the density of logical knowledge by aggregating the topological complexity of the context and the derivation path. This measurement encompasses contextual logical density for premise structures, inferential topology density for inferential depth, and SLKD score aggregation for synthesizing these dimensions into a unified metric.

\subsubsection{Contextual Logical Density}
First, we measure the static logical information contained in the problem description. The Context Density $S_{\text{ctx}}$ is calculated as:

\begin{equation}
S_{\mathrm{ctx}}=\underbrace{|{E}_i|\cdot\bar{D}^2}_{\text{Structural}}+\underbrace{|{P}_i|}_{\mathrm{Predicate}}+\underbrace{|{C}_i|}_{\mathrm{Constant}}
\end{equation}

where $|E_i|$ counts logical expressions, $\bar{D}$ is their average nesting depth (calculated via parse tree analysis), $|P_i|$ and $|C_i|$ tally unique predicates and constants respectively.

Here, the term $\bar{D}^2$ applies a quadratic penalty to the nesting depth, reflecting the non-linear increase in cognitive load required to encode deeply nested logical structures. This formulation ensures that complex conditional dependencies contribute significantly more to the density score than simple linear facts.

\subsubsection{Inferential Topology Density} 
Next, we quantify the dynamic complexity of the entailment path required to validate a specific option. For an option $l$, its inferential topology  density $S^{(l)}_{\text{opt}}$ is defined as:

\begin{equation}
S_{\mathrm{opt}}^{(l)}=\underbrace{|{R}_l|\cdot\bar{D}_l^2}_{\text{Preconditions}}+\sum_{k=1}^{T_l}\underbrace{\left(1+\#\text{Operations}_{l,k}\right)D_{l,k}^2}_{\mathrm{Step~}k}
\end{equation}

where $l \in \{1,...,\text{L}\}$ indexes the \text{L} answer options, $R_l$ is the set of precondition expressions for
option $l$ with average nesting depth $\bar{D_l}$, and $T_l$ is the number of reasoning steps for option l. Each 
step $k$ contributes according to its operator count $\#\text{Operations}_{l,k}$, and nesting depth $D_{l,k}$.



This component aggregates the structural weight of each derivation step. By summing over the sequence $T_l$, we capture the total "logical work" encoded in the reasoning chain.

\subsubsection{\ours Score Aggregation}
The final \ours $S$ is obtained by combining the contextual and derivational densities. To facilitate its use as a weighting factor in training, we normalize the raw score into a probabilistic range $[0, 1]$:
\begin{equation}
    S_{\text{raw}} = S_{\text{ctx}} + \sum_{l=1}^{L} S^{(l)}_{\text{opt}}
\end{equation}

\begin{equation}
    S = \sigma \left( \gamma \cdot \frac{\log(S_{\text{raw}} + 1) - \mu}{\sqrt{\delta
    ^2 + \epsilon}} + \beta \right)
\end{equation}

where $\sigma$ denotes the sigmoid function, $\mu$ and $\delta$ are dataset statistics, and parameters ($\gamma = 1$,$\beta = 0$,$\epsilon = 10^{-5}$) ensure stable $[0,1]$ normalization.

This normalized \ours score serves as a proxy for the richness of reasoning patterns contained in a sample, independent of the sample's surface length.

\begin{figure*}[hbt!]
    \centering
    \includegraphics[width=1.0\textwidth]{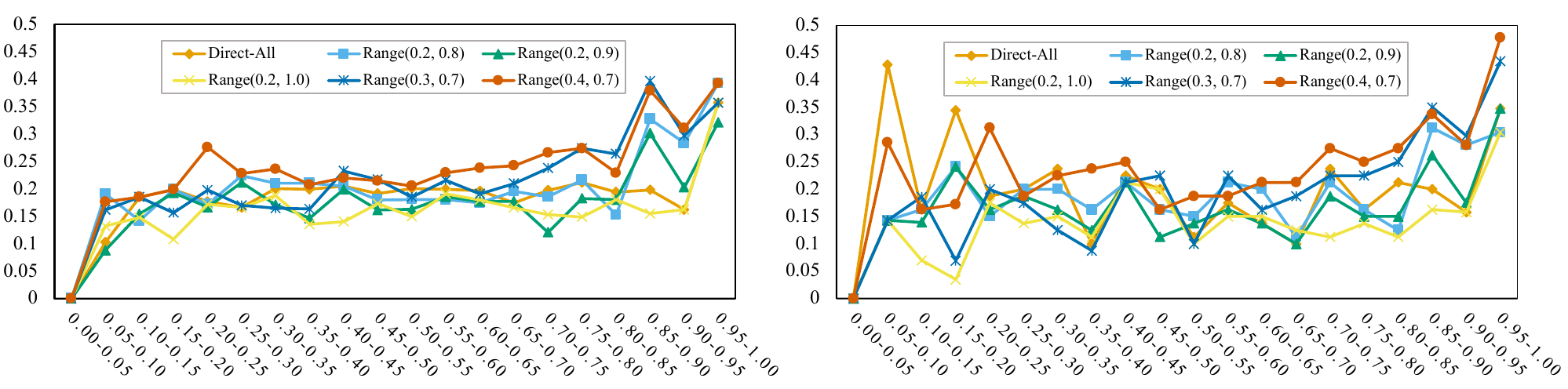}
    \caption{Experimental results of fine-tuning models in different intervals. \textit{"Direct-All"} denotes a model fine-tuned on all training examples. \textit{"Range(x, y)"} denotes a model fine-tuned on examples whose \ours scores fall between $x$ and $y$.
    Left: Test dataset experiment results.
    Right: Balanced testing set experiment results.    
    The horizontal axis represents score intervals, and the vertical axis represents error rates (lower error indicates better performance).}
    \label{fig:3_Original distribution}
    \vspace{-4mm}
\end{figure*}

\begin{figure*}[hbt!]
    \centering
    \includegraphics[width=1.0\textwidth]{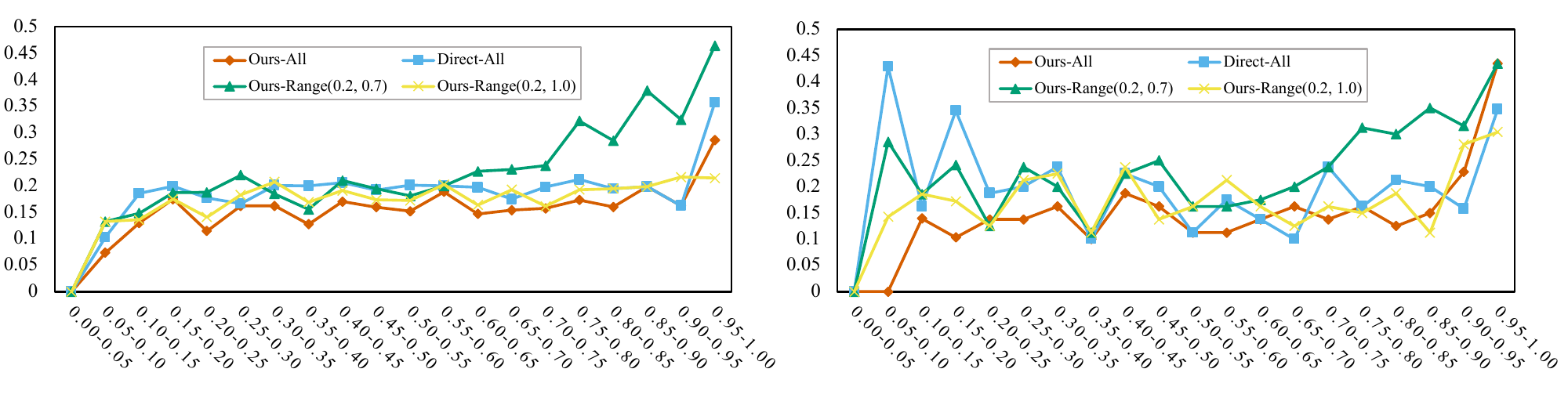}
    \caption{Experimental results of fine-tuning models using different methods. \textit{"Ours-All"}: model fine-tuned on the full training set using \textit{re-cognizing optimization}. \textit{"Ours-Range(x, y)"}: model fine-tuned with \textit{re-cognizing optimization} on examples whose \ours scores fall between $x$ and $y$.
    Left: Test dataset results.
    Right: Balanced testset results.
    The horizontal axis represents score intervals and the vertical axis represents error rates (lower error indicates better performance).}
    \label{fig:4_experimental_analysis}
    \vspace{-4mm}
\end{figure*}

\section{Empirical Analysis for \ours}
In order to fully investigate the effect of \ours, we conduct experiments in existing training data (experimental setting is in Appendix~\ref{appendix:exp_details}).
We not only focus on the \ours score distribution of the existing data but also use data with different interval distributions as training sets to separately validate the effectiveness of the data within these intervals.


\subsection{\ours Score Distribution} 
In Figure~\ref{fig:Data_Distribution_2}(a), when the training set is binned into 20 score intervals, the sample counts form a bell-shaped curve with a weighted mean of $\mu = 0.526$ and standard deviation $\sigma = 0.204$. This approximately Gaussian frequency distribution demonstrates that our \ours scores effectively separate examples by their underlying reasoning potential. Figure~\ref{fig:Data_Distribution_2}(a) also plots model error rate against these intervals: error climbs from 8.5\% at a score of 0.1 to 59.3\% at 0.6, demonstrating a clear positive correlation between \ours and failure rate. Beyond a score of 0.6, error rate plateaus at approximately 61\% ± 3.6\%. This saturation likely arises because (1) very high \ours items exceed the model’s current reasoning capacity, and (2) the top bins contain few, uniformly hard cases, so failures become uniformly pervasive. Unlike surface metrics (e.g., sentence length or vocabulary complexity), our score distribution captures deeper logical structure.

\subsection{\ours Score Distribution Balancing} 
The original testset contains 4,743 examples and its \ours score distribution (Figure~\ref{fig:Data_Distribution_2}(b)) reveals that 74.3\% of samples fall into the mid-range interval \texttt{(0.2, 0.7)}, while only 17.2\% lie in the high \ours interval \texttt{(>0.7)}.
Such skew can introduce two evaluation biases:
(1) Apparent overfitting in the mid-range: strong results in the overrepresented middle interval may conceal weaknesses on higher \ours examples;
(2) Insufficient statistical power: the small number of high \ours samples ($n = 818$) yields wide confidence intervals for any observed gains.
To correct this, \textbf{we build a balanced benchmark} (Figure~\ref{fig:Data_Distribution_2}(c)) by drawing 80 examples from each interval.
This uniform sampling ensures equal representation across the \ours spectrum, eliminating bias and enabling reliable comparisons.

\subsection{Data Potential Analysis by \ours}
Building on our validation of the \ours score and the creation of a balanced test benchmark, we next investigated how filtering training examples by their \ours scores affects learning.
In Figure~\ref{fig:3_Original distribution}, we systematically evaluate model performance when trained on subsets defined by \ours intervals, using both the original and balanced test sets. These experiments reveal three critical patterns, which can be summarized as the following observations:

\begin{itemize}[
    label=\raisebox{0.25ex}{\scalebox{0.6}{$\bullet$}},
    leftmargin=1em,   
    itemindent=0pt,   
    noitemsep,        
    topsep=0pt        
]
    \item \textbf{Obs 1}: \textbf{\textit{Low \ours data can be safely pruned.}}
    Both Figure~\ref{fig:3_Original distribution}(a) and (b) show that training on \texttt{Range(0.2, 1.0)}, which omits only the lowest 20\% of examples, consistently outperforms full-data training, reducing error rates in nearly every bin. Even when further restricting to \texttt{Range(0.2, 0.8)} (removing both the lowest and highest 20\%), overall accuracy remains on par with or slightly above the full-data baseline ($\Delta = +0.3\%$), and mid-range bins \texttt{(0.2, 0.7)} see notable gains. These results confirm that low \ours examples can be pruned without harming model performance and can often lead to improvements, while indiscriminate use of all data may introduce noise.

    \item \textbf{Obs 2}: \textbf{\textit{High \ours data are catalysts for improvement.}}  
    Figure~\ref{fig:3_Original distribution}(a) shows that training on \texttt{Range(0.2, 1.0)} yields the lowest error rates across all bins. When the upper bound is reduced to \texttt{Range(0.2, 0.9)}, error rates rise, particularly in the highest \ours bins. Narrowing further to \texttt{Range(0.2, 0.8)} causes a further increase in errors. This stepwise degradation, also reflected in Figure~\ref{fig:3_Original distribution}(b), confirms that examples with the highest \ours scores drive the most significant performance gains and act as catalysts for model learning.

    \item \textbf{Obs 3}: \textbf{\textit{Too little data breaks the learning process.}}  
    Figures~\ref{fig:3_Original distribution}(a) and (b) show that restricting training to narrow \ours intervals \texttt{Range(0.3, 0.7)} or \texttt{Range(0.4, 0.7)} leads to severe performance degradation, especially in the mid \ours range \texttt{(0.2, 0.7)}, where error rates exceed those from broader training ranges. This demonstrates that sparse coverage of the \ours spectrum impairs the model’s ability to internalize and apply core reasoning patterns. The sharp drop in performance underscores the necessity of preserving sufficient data diversity and quantity across all \ours levels to sustain learning.
\end{itemize}

In summary, these results indicate that \textbf{a large volume of data only provides sparse logical knowledge, while the importance of samples with high \ours has been overlooked.}
To address these issues, we propose the \textit{re-cognizing optimization} method in Section \ref{sec:optimization}.

\section{\ours-based Re-cognizing Method}



\label{sec:optimization}
Inspired by cognitive theories \citep{Cognitive_Load_Sweller, Resource_rational_Lieder}, we propose a two-phase \textit{re-cognizing optimization}. 
Our optimization method is guided by \ours and divided into two phases.
We first recalibrates the LLM’s logical reasoning schema and then leverages \ours scores to reinforce its logical reasoning pathways.

\subsection{{Phase I: Model Cognition Reshaping}}  
In this phase, we reorder the training data according to \ours scores to "reset" and align the model's reasoning framework, applying Sweller’s cognitive load theory. Allowing the model to explore the full spectrum of \ours examples from the outset helps it form broad reasoning patterns. This “low‐stakes exploration” mirrors how human learners build foundational knowledge before tackling more challenging tasks, minimizing extraneous cognitive load and establishing a robust framework for subsequent learning.

\subsection{{Phase II: Cognitive Reasoning Enhancement}}  
Here, we implement resource-rational analysis by guiding the model’s focus according to normalized \ours scores. The probability $p$ that the model attends to sample $i$ is calculated as follows:
\begin{equation}
p_i \;=\;\frac{\hat s_i}{\sum_{j=1}^N \hat s_j},
\quad
\hat s_i = \frac{s_i - s_{\min}}{s_{\max} - s_{\min}}
\end{equation}
where $s_i$ is the raw \ours score of sample $i$, $s_{\min}$ and $s_{\max}$ are the minimum and maximum scores in the dataset, $\hat s_i$ is the normalized score for sample $i$, and the denominator $\sum_{j=1}^N \hat s_j$ sums these normalized scores over all $N$ samples.
The \textit{re-cognizing optimization} stratege echoes Lieder’s insight that "human cognition allocates limited resources to maximize expected reasoning gains relative to reasoning demands". 
Through cognition‐driven emphasis, the model is steered toward high \ours examples, thus securing greater learning returns under constrained resources.

\begin{table*}[ht!]
\centering
\footnotesize
\resizebox{\textwidth}{!}{
\begin{tabular}{>{\raggedright\arraybackslash}p{1.5cm}lcccccccccc} 
\toprule

\multirow{2}{1.5cm}{\textsc{Model}} & \multirow{2}{*}{\textsc{Methods}} & \multicolumn{5}{c}{\textsc{Unbalanced}} & \multicolumn{5}{c}{\textsc{Balanced}}  \\
\cmidrule(lr){3-7} \cmidrule(lr){8-12}
& & \multicolumn{1}{c}{\fontsize{8pt}{6pt}\selectfont Recolr}   
& \multicolumn{1}{c}{\fontsize{8pt}{6pt}\selectfont LogiQA}  
& \multicolumn{1}{c}{\fontsize{8pt}{6pt}\selectfont LogiQA2.0} 
& \multicolumn{1}{c}{\fontsize{8pt}{6pt}\selectfont LogicBench} 
& \multicolumn{1}{c}{\fontsize{8pt}{6pt}\selectfont Avg.}  
& \multicolumn{1}{c}{\fontsize{8pt}{6pt}\selectfont Recolr}   
& \multicolumn{1}{c}{\fontsize{8pt}{6pt}\selectfont LogiQA} 
& \multicolumn{1}{c}{\fontsize{8pt}{6pt}\selectfont LogiQA2.0} 
& \multicolumn{1}{c}{\fontsize{8pt}{6pt}\selectfont LogicBench} 
& \multicolumn{1}{c}{\fontsize{8pt}{6pt}\selectfont Avg.} \\

\midrule

\multirow{2}{1.5cm}{\scriptsize \textit{Closed-source LLMs}} & \scriptsize GPT-4 & \scriptsize 0.808  & \scriptsize 0.525  & \scriptsize 0.664  & \scriptsize 0.774  & \scriptsize 0.707  & \scriptsize 0.794  &\scriptsize 0.588  & \scriptsize 0.651  & \scriptsize 0.765  & \scriptsize 0.699 \\
& \scriptsize DeepSeek-V3 & \scriptsize 0.754  & \scriptsize 0.484  & \scriptsize 0.663  & \scriptsize 0.814  & \scriptsize 0.712  & \scriptsize 0.738  & \scriptsize 0.500  & \scriptsize 0.663  & \scriptsize 0.847  & \scriptsize 0.686 \\

\midrule

\multirow{8}{1.5cm}{} & \scriptsize Base & \scriptsize 0.444 & \scriptsize 0.347  & \scriptsize 0.378   & \scriptsize 0.707   & \scriptsize 0.521  & \scriptsize 0.447 & \scriptsize 0.412 & \scriptsize 0.383  & \scriptsize 0.714  & \scriptsize 0.490 \\

& \scriptsize Directly & \scriptsize 0.896 & \scriptsize 0.724 & \scriptsize 0.805  & \scriptsize 0.812  & \scriptsize 0.806  & \scriptsize 0.901 & \scriptsize 0.774  & \scriptsize 0.802  &  \scriptsize 0.812  & \scriptsize 0.816 \\

& \scriptsize Curriculum Learning & \scriptsize 0.926 & \scriptsize 0.634 & \scriptsize 0.798  & \scriptsize 0.854  & \scriptsize 0.811  & \scriptsize 0.908 & \scriptsize 0.684  & \scriptsize 0.773  &  \scriptsize 0.864  & \scriptsize 0.815 \\

& \scriptsize Bin-based Progressive Learning & \scriptsize 0.930 & \scriptsize 0.742 & \scriptsize 0.821  & \scriptsize 0.831  & \scriptsize 0.826  & \scriptsize 0.894 & \scriptsize 0.763  & \scriptsize 0.809  &  \scriptsize 0.836  & \scriptsize 0.823 \\

\rowcolor{gray!15}
\cellcolor{white}
& \scriptsize  \textit{Optimization w/o Stage1} & \scriptsize 0.904 & \scriptsize 0.699 & \scriptsize 0.791  & \scriptsize 0.747  & \scriptsize 0.771  & \scriptsize 0.918 & \scriptsize 0.726  & \scriptsize 0.825  &  \scriptsize 0.763  & \scriptsize 0.806 \\

\rowcolor{gray!15}
\cellcolor{white}
& \scriptsize \textit{Optimization w/o Stage2} & \scriptsize 0.910 & \scriptsize 0.717 & \scriptsize 0.796  & \scriptsize 0.824  & \scriptsize 0.809  & \scriptsize 0.918 & \scriptsize 0.768  & \scriptsize 0.834  &  \scriptsize 0.859  & \scriptsize 0.844 \\

\rowcolor{gray!15}

\cellcolor{white}
& \scriptsize \textit{Our Re-Cognizing Method} & \scriptsize \textbf{0.930} & \scriptsize \textbf{0.750} & \scriptsize \textbf{0.835}   & \scriptsize \textbf{0.857}   & \scriptsize \textbf{0.843}  & \scriptsize \textbf{0.922} & \scriptsize \textbf{0.775}  & \scriptsize \textbf{0.840}  &  \scriptsize \textbf{0.865}  & \scriptsize \textbf{0.851}\\

\cellcolor{white} \multirow{-8}{1.5cm}{\scriptsize LLaMA3.1-8B} & \scriptsize \textit{(relative gain)} 
& \scriptsize \color{ForestGreen}{+0.00\%} & \scriptsize \color{ForestGreen}{+1.08\%} & \scriptsize \color{ForestGreen}{+1.71\%} & \scriptsize \color{ForestGreen}{+0.35\%} & \scriptsize \color{ForestGreen}{+2.06\%} 
& \scriptsize \color{ForestGreen}{+1.54\%} & \scriptsize \color{ForestGreen}{+0.13\%} & \scriptsize \color{ForestGreen}{+3.83\%} & \scriptsize \color{ForestGreen}{+0.12\%} & \scriptsize \color{ForestGreen}{+3.40\%} \\

\midrule

\multirow{8}{1.5cm}{} & \scriptsize Base & \scriptsize 0.466 & \scriptsize 0.372  & \scriptsize 0.424   & \scriptsize 0.629   & \scriptsize 0.509  & \scriptsize 0.433 & \scriptsize 0.435 & \scriptsize 0.409  & \scriptsize 0.678  & \scriptsize 0.490 \\

& \scriptsize Directly & \scriptsize 0.786 & \scriptsize 0.743 & \scriptsize 0.748  & \scriptsize 0.810  & \scriptsize 0.778  & \scriptsize 0.797 & \scriptsize 0.719  & \scriptsize 0.788  &  \scriptsize 0.888  & \scriptsize 0.798 \\

& \scriptsize Curriculum Learning & \scriptsize 0.864 & \scriptsize 0.750 & \scriptsize 0.806  & \scriptsize 0.850  & \scriptsize 0.823  & \scriptsize 0.872 & \scriptsize 0.739  & \scriptsize 0.797  &  \scriptsize 0.919  & \scriptsize 0.831 \\

& \scriptsize Bin-based Progressive Learning & \scriptsize 0.890 & \scriptsize 0.730 & \scriptsize 0.782  & \scriptsize 0.842  & \scriptsize 0.812  & \scriptsize 0.911 & \scriptsize 0.697  & \scriptsize 0.806  &  \scriptsize 0.719  & \scriptsize 0.833 \\

\rowcolor{gray!15}
\cellcolor{white}
& \scriptsize  \textit{Optimization w/o Stage1} & \scriptsize 0.856 & \scriptsize 0.808 & \scriptsize 0.817  & \scriptsize 0.833  & \scriptsize 0.827  & \scriptsize 0.833 & \scriptsize 0.790  & \scriptsize 0.822  &  \scriptsize 0.894  & \scriptsize 0.835 \\

\rowcolor{gray!15}
\cellcolor{white}
& \scriptsize  \textit{Optimization w/o Stage2} & \scriptsize 0.798 & \scriptsize 0.679 & \scriptsize 0.767  & \scriptsize 0.827  & \scriptsize 0.784  & \scriptsize 0.839 & \scriptsize 0.684  & \scriptsize 0.778  &  \scriptsize 0.872  & \scriptsize 0.793 \\

\rowcolor{gray!15}
\cellcolor{white}
& \scriptsize \textit{Our Re-Cognizing Method} & \scriptsize \textbf{0.944} & \scriptsize \textbf{0.824} & \scriptsize \textbf{0.845}   & \scriptsize \textbf{0.858}   & \scriptsize \textbf{0.858}  & \scriptsize \textbf{0.948} & \scriptsize \textbf{0.797}  & \scriptsize \textbf{0.884}  &  \scriptsize \textbf{0.919}  & \scriptsize \textbf{0.886}\\

\cellcolor{white} \multirow{-8}{1.5cm}{\scriptsize Qwen2.5-7B} & \scriptsize \textit{(relative gain)} 
& \scriptsize \color{ForestGreen}{+6.07\%} & \scriptsize \color{ForestGreen}{+9.87\%} & \scriptsize \color{ForestGreen}{+4.84\%} & \scriptsize \color{ForestGreen}{+0.94\%} & \scriptsize \color{ForestGreen}{+4.25\%} 
& \scriptsize \color{ForestGreen}{+4.06\%} & \scriptsize \color{ForestGreen}{+7.85\%} & \scriptsize \color{ForestGreen}{+9.68\%} & \scriptsize \color{ForestGreen}{+0.00\%} & \scriptsize \color{ForestGreen}{+6.36\%} \\

\bottomrule
\end{tabular}}
\caption{Experimental results from different test sets. Our \textit{re-cognizing optimization} method is compared against other approaches on both LLaMA3.1-8B and Qwen2.5-7B using accuracy as the evaluation metric. Results are reported on both unbalanced and balanced test sets, with the best performance in each setting highlighted in bold.}
\label{tab:main}
\vspace{-4mm}
\end{table*}

\subsection{Results and Analysis }
\begin{figure*}[hbt!]
    \centering
    \includegraphics[width=1.0\textwidth]{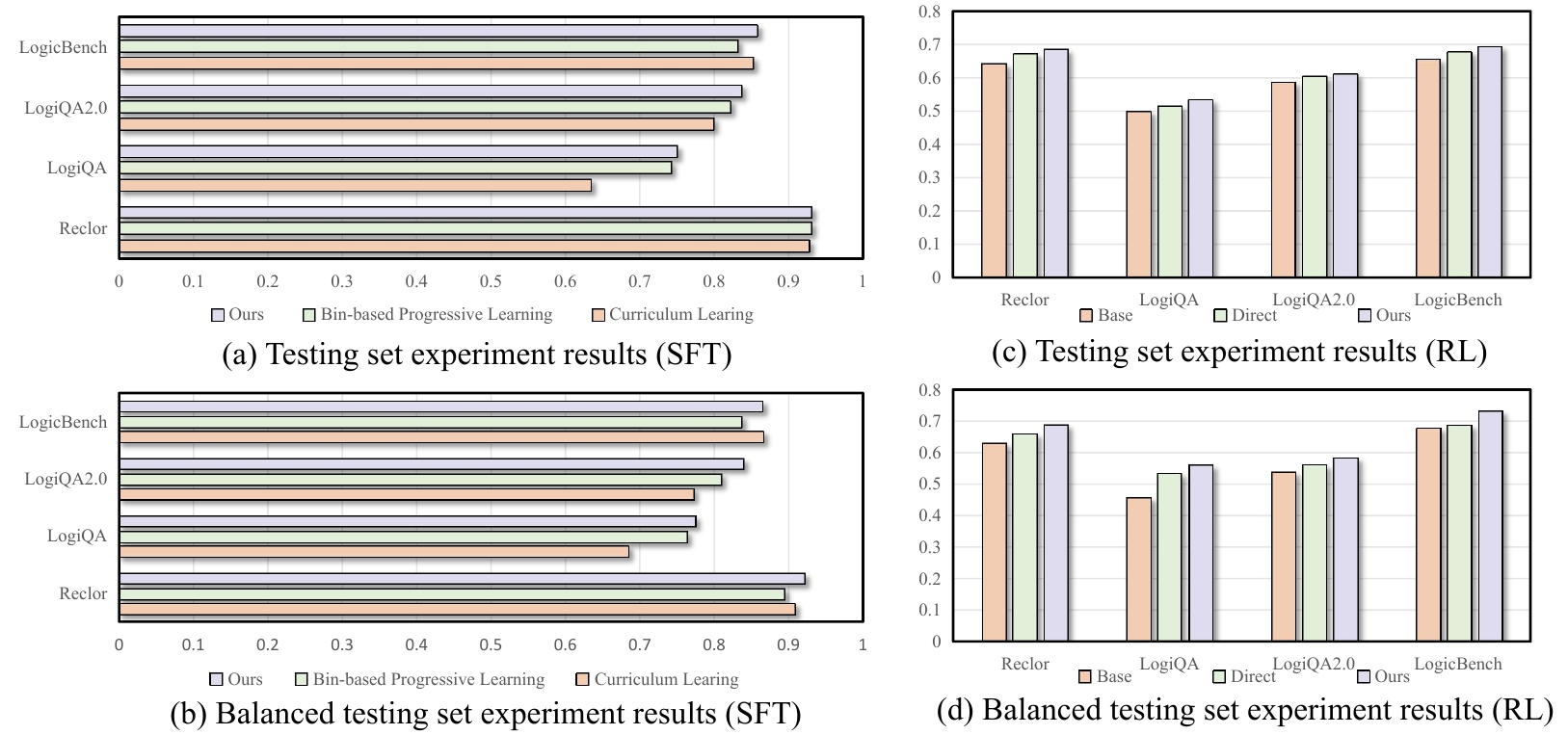}
    \caption{Experimental results of SFT and RL across different dataset. Accuracy is used as the evaluation metric (higher is better). Both SFT and RL methods are evaluated on the original and balanced test sets. (a) SFT on the original test set: Ours vs. Bin-based Progressive Learning vs. Curriculum Learning. (b) SFT on the balanced test set: same comparisons. (c) RL on the original test set: Ours vs. Base vs. Direct static reward). (d) RL on the balanced test set: same comparisons. In every setting, our method (Ours) achieves the highest accuracy, demonstrating its superiority under both SFT and RL regimes.}
    \label{fig:5_Expansion}
    \vspace{-4mm}
\end{figure*}

We evaluate the effectiveness of the \textit{re-cognizing optimization} strategy by comparing it with static baselines and other data-centric methods across datasets. Other methods include curriculum learning \citep{CL_base_paper} and bin-based progressive learning \citep{Bin-based_Progressive_learning}. In curriculum learning, training examples are introduced in order of increasing \ours scores; in bin-based progressive learning, data is divided into different \ours bins, and the next \ours bin is introduced only after the model has been trained on the current bin. Table~\ref{tab:main} summarizes the performance across datasets, while Figure~\ref{fig:4_experimental_analysis} shows the error rates across \ours bins for the original and balanced test sets. Figure~\ref{fig:5_Expansion} (left) compares our method with the other two methods. From the experimental results, we have the following observations:

\begin{itemize}[
    label=\raisebox{0.25ex}{\scalebox{0.6}{$\bullet$}},
    leftmargin=1em,   
    itemindent=0pt,   
    noitemsep,        
    topsep=0pt        
]
    \item \textbf{Obs 1:} \textbf{\textit{Re-cognizing optimization consistently reduces errors across datasets and \ours bins, leading to comprehensive and robust performance gains.}}  
    In Figure~\ref{fig:4_experimental_analysis}, Figure~\ref{fig:5_Expansion} (left), and Table~\ref{tab:main}, our method consistently achieves the best performance across datasets and test splits, confirming its ability to generalize dataset-specific reasoning patterns. For example, in the lowest \ours interval \texttt{(0, 0.2)}, errors drop from 17.6\%/25\% to 12.6\%/11.2\% on the original/balanced test sets, demonstrating reduced overfitting to trivial cases. This trend holds across all \ours bins, resulting in a systematic reduction in errors across the spectrum. On datasets like LogiQA and LogiQA2.0, where baseline accuracies are lower, our method shows a significant improvement over others, demonstrating its adaptability and robustness on diverse datasets.
 
    \item \textbf{Obs 2:} \textbf{\textit{Re-cognizing optimization demonstrates resilience in data-scarce settings by maintaining strong performance even with restricted training ranges.}}  
    In Figure~\ref{fig:4_experimental_analysis}, when we apply \textit{re-cognizing optimization} to restricted training sets—\texttt{Our-Range(0.2, 0.7)} and \texttt{Our-Range(0.2, 1.0)}—we observe a 13.6\% error increase when high \ours examples (\textgreater 0.7) are omitted. However, our method effectively compensates for this loss, maintaining performance comparable to full-data training within the \texttt{Our-Range(0.2, 0.7)} subset ($\Delta = +2.5\%$).
    Expanding the training range to include higher \ours examples (\texttt{Our-Range(0.2, 1.0)}) consistently maintains lower error rates than direct full-data training, validating the predictive power of our \ours and demonstrating that our method can effectively improve performance even in data-scarce settings.
\end{itemize}

\paragraph{Ablation Study}
To assess the contribution of each training stage, we conduct ablation experiments by removing either Stage 1 or Stage 2 from the full \textit{re-cognizing optimization} pipeline. As shown in Table~\ref{tab:main}, both variants lead to substantial performance drops on both LLaMA3.1-8B and Qwen2.5-7B. This confirms that both stages are essential for achieving optimal reasoning performance.
In summary, our results show our \textit{re-cognizing optimization} effectively directs the model to high \ours examples, boosting learning efficiency. It systematically reduces errors across \ours bins (proving improved reasoning) and outperforms traditional methods like curriculum and bin-based learning, demonstrating flexibility and robustness across scenarios.

\subsection{Expansion Exploration}
Some studies \citep{vl-thinking2025, DBLP:journals/corr/abs-2501-17161,RL_data_selector} suggest that supervised fine-tuning (SFT) and reinforcement learning (RL) play different roles in stimulating the capabilities of language models. 
To assess the impact of our \ours scores in an RL setting,
we provide proportionally larger accuracy rewards for samples with higher \ours scores, incentivizing the model to handle more reasoning-intensive cases and internalize richer reasoning patterns. We also enforce a structured output format and reward the model for both format compliance and the quality of its reasoning trace. Figure~\ref{fig:5_Expansion}(c) and (d) compare our method against two baselines: the base model without RL and a direct variant with fixed accuracy rewards. In both the original and balanced test sets, our \textbf{SLKD-guided} method consistently outperforms these baselines, confirming that dynamic, score-based rewards combined with structured output incentives significantly enhance the model’s reasoning capabilities.

\section{Discussion}

Our results resonate with the \textit{Densing Law of LLMs} \citep{xiao2024densinglawllms}, which observes exponential growth in model "capability density".
By positing that \textit{Logical Structure is Knowledge}, we extend this intuition to the data dimension. This aligns with the view of language modeling as data compression \citep{DBLP:conf/iclr/DeletangRDCGMGW24}, suggesting that reasoning potential depends less on token volume and more on the concentration of logical primitives—a principle validated by the Phi-series \citep{DBLP:journals/corr/abs-2404-14219}. We thus advocate for a shift from volume-based Scaling Laws \citep{DBLP:journals/corr/abs-2001-08361} toward a structural density-aware paradigm.

Regarding a universal data knowledge density, our analysis suggests that density is likely a relative metric tied to LLMs’ cognitive boundary rather than an absolute constant.
The gains from high-\ours samples highlight a logical sparsity in current corpora, where redundant patterns dilute learning signals \citep{abbas2023semdedupdataefficientlearningwebscale}.
Future work should investigate whether excessive logical density triggers distribution shifts or overfitting to formal structures, particularly when interacting with LLMs' internal mechanisms and inductive biases.

\section{Conclusion}

In this paper, we presented a  data-centric framework  to bolster the reasoning capabilities of LLMs. 
By introducing \textit{Structured Logical Knowledge Density}  (\ours), we provide a formal metric to quantify the intrinsic logical complexity of training samples.
Furthermore, our proposed re-cognizing optimization strategy effectively aligns training data with the inherent reasoning boundaries of LLMs. 
Extensive experiments  demonstrate that our approach consistently outperforms existing baseline methods. 
We hope our findings offer a fresh perspective on measuring and unlocking the untapped reasoning potential of LLMs, while inspiring future research into cognitive-level data engineering.

\newpage

\section*{Limitation}
\label{append:Limitation}
One limitation of our work is the relatively small scale and variety of models we evaluated—due to budget and time constraints, we focused on only two model families (e.g., LLaMA3.1-8B and Qwen2.5-7B), which may limit the generalizability of our findings to larger or more diverse architectures. Additionally, while our \ours score provides a useful proxy for reasoning potential, its formulation could be refined further: the current metrics may not capture all aspects of reasoning complexity or transfer seamlessly to other task domains. 
Moreover, our logical element extraction relies on LLM-based distillation functions, which can introduce noise or inaccuracies; nevertheless, we found these errors to be minor and within acceptable bounds, having minimal impact on overall metric reliability. 
Finally, our use of GRPO-based reinforcement learning was exploratory and preliminary; more extensive experiments with alternative reward schemes, longer training runs, and varied model capacities will be necessary to fully assess the robustness and scalability of \textit{re-cognizing optimization}. 
We also did not explore individualized learning trajectories emphasized in cognitive science, nor use our reasoning-intensity signal to dynamically switch between System 1 and System 2 modes.
Moreover, we restrict our evaluation to fine-tuning existing pretrained checkpoints rather than full-scale pretraining, since retraining multi-billion-parameter models from scratch requires resources beyond our current capabilities.

\bibliography{acl_latex}



\appendix

\newpage

\section{Impact, LLM Usage and Future Work}

We acknowledge that although our \ours score and \textit{re-cognizing optimization} method are effective for several tasks in our research, they are far from being perfect. Here, we honestly discuss the limitations, broader impact, and potential avenues for future works.

\subsection{The Use of Large Language Models}
\label{llm_use}
We hereby state that LLMs were deployed purely for translation and linguistic refinement tasks. All research ideas, experimental protocols, execution processes, analytical work, and derived conclusions are the exclusive responsibility of the authors. A comprehensive validation of the manuscript has been performed to guarantee that the employment of LLMs did not introduce any false or misleading information.

\subsection{Impact}
Our \ours score is a pioneering metric that quantifies each example’s reasoning potential. Paired with our \textit{re-cognizing optimization} framework, which uses this score to guide training, our approach reduces unnecessary computation, boosts model reasoning performance, and supports more sustainable AI practices. We believe this work will inspire new directions and offer systematic guidance for future research on unlocking the latent potential of LLM training data. Societally, our method can positively influence the efficient training of LLMs, enabling the creation of more robust and resource-efficient AI systems.

\subsection{Future Work}  
\textbf{Broader model and task coverage.} 
In future work, it would be valuable to evaluate our framework on additional architectures—beyond LLaMA-7B and Qwen2.5-7B—and across new domains such as mathematical reasoning, code generation, and multimodal understanding. This broader testing would help establish the generality and limits of \textit{Structured Logical Knowledge Density} and \textit{re-cognizing optimization}.

\textbf{Refined reinforcement learning integration.} 
While we have already incorporated reasoning-intensity scores into GRPO rewards, it would be useful to explore more sophisticated applications—such as dynamic reward shaping, alternative RL algorithms, or multi-objective formulations—to further boost learning efficiency and stability.

\textbf{Adaptive reasoning routing.} 
We also plan to investigate using reasoning-intensity as a runtime signal to guide the model’s choice of reasoning mode, enabling dynamic switching between fast, heuristic processing (System 1) and deeper, deliberative reasoning (System 2). This may prevent overthinking on trivial inputs and ensure adequate effort on challenging ones.

\textbf{Compatibility with alternative data-selection methods.}  
Our \textit{Structured Logical Knowledge Density} metric and \textit{re-cognizing optimization} are fully compatible with other sampling strategies. In future work, we will experiment with hybrid schemes that combine our method with these complementary approaches to maximize sample efficiency and model performance and to provide a more rigorous comparison against established data-selection techniques.

\textbf{Pretraining-stage integration.}
An exciting direction is to extend reasoning-intensity guidance into the pretraining curriculum. Although pretraining multi-billion-parameter models from scratch exceeds our current resources, future work could integrate \ours into early model training to shape core reasoning capabilities, contingent on access to sufficient compute.

\begin{algorithm}[ht!]
    \caption{\ours Calculation}
    \label{alg:scoring}
    \begin{algorithmic}[1] 
        \State \textbf{Input:} A sample $x$ with context $c$ and options $\{o_l\}_{l=1}^L$.
        \State Parse context $c$:
        \State \quad Extract logical expressions E and compute their nesting depths.
        \State \quad Identify predicates ${P}$ and constants ${C}$ in $c$.
        \State Compute context intensity:
        \State \quad $S_{\text{ctx}} = |{E}| \times \bar{D}^2 + |{P}| + |{C}|$.
        \For{\textbf{each} option $o_l$}
            \State Extract preconditions ${R}_l$ and reasoning steps ${S}_l$.
            \State Compute option intensity:
            \State $S_{\text{opt}}^{(l)} \;=\;
|{R}_l| \cdot \bar{D}_l^2
\;+\;\sum_{k=1}^{T_l}{\bigl(1+\#\text{Operations}_{l,k}\bigr)\;D_{l,k}^2}$.
        \EndFor
        \State Aggregate raw intensity:
        \State \quad $S_{\text{raw}} = S_{\text{ctx}} + \sum_{l=1}^L S_{\text{opt}}^{(l)}$.
        \State Normalize to $[0,1]$ via sigmoid of log:
        \State $
        \begin{aligned}
        S &= \sigma\left(\gamma \cdot \frac{\log(S_{\text{raw}}+1) - \mu}{\sqrt{\delta^2+\epsilon}} + \beta\right)
        \end{aligned}.
        $
        \State \textbf{Output:} reasoning-intensity score $S$.
    \end{algorithmic}
\end{algorithm}

\begin{algorithm}[ht!]
    \caption{Re-Cognizing Optimization}
    \label{alg:dws}
    \begin{algorithmic}[1]
        \State \textbf{Input:} dataset ${D}=\{x_i\}_{i=1}^N$, model $M$, epochs $T$.
        \State Precompute intensity scores $\{S_i\}_{i=1}^N$ via Algorithm~\ref{alg:scoring}.
        \For{epoch $t = 1$ \textbf{to} $T$}
            \If{$t = 1$}
                \State Uniformly shuffle ${D}$ for initial exploration (Phase I: Model Cognition Reshaping).
            \Else
                \State Sort ${D}$ by descending $S_i$ to emphasize high-intensity samples (Phase II: Cognitive Reasoning Enhancement).
            \EndIf
            \For{each batch $B$ drawn sequentially from ${D}$}
                \State Compute loss on $B$ and update model parameters.
            \EndFor
        \EndFor
        \State \textbf{Output:} fine-tuned model $M^*$.
    \end{algorithmic}
\end{algorithm}

\section{\ours Calculation Process and Re-Cognizing Optimization Algorithm}
\label{append:code}
Algorithm 1 illustrates the calculation process of our \ours score, and Algorithm 2 demonstrates the implementation procedure of our Re-Cognizing Optimization strategy.

\section{Details for the  Experiments}\label{sec:details_experiments}
\label{appendix:exp_details}

\subsection{Experimental Settings}
In this section, we conduct experiments on four logical reasoning benchmarks: \textit{Reclor} (contextual reasoning) \citep{Reclor}, \textit{LogicBench} (deductive reasoning) \citep{LogicBench}, \textit{LogiQA} (workplace logic analysis) \citep{LogiQA}, and \textit{LogiQA2.0} (enhanced adversarial patterns) \citep{LogiQA2.0}.
For our experiments, we utilize two backbone models: \textit{LLaMA3.1-8B-Instruction} \citep{llama3} and \textit{Qwen2.5-7B-Instruction} \citep{Qwen2.5}. We refer to them hereafter as \textit{LLaMA3.1-8B} and \textit{Qwen2.5-7B}. Additionally, we include GPT-4 \citep{gpt4-technical-report} and DeepSeek-V3 \citep{deepseekv3-technical-report} as reference models. For more implementation details, please refer to the Appendix \ref{sec:details_experiments}.


The released code for this work is available at 
\url{https://github.com/HuzhouNLP/Logical_Structure_as_Knowledge}.

\subsection{Training Configuration}
Our datasets consist of 36,788 training instances and 4,743 test instances, covering a broad range of logical challenges, including propositional logic, syllogisms, and temporal reasoning.

Both \textit{LLaMA3.1-8B} and \textit{Qwen2.5-7B} are fine-tuned using supervised fine-tuning (SFT) under the TRL framework, employing LoRA modules with a rank of 64, $\alpha=16$, and a dropout rate of 0.05. In addition to the SFT experiments, we also conduct reinforcement learning (RL) experiments on \textit{Qwen2.5-7B}. For RL, we apply the GRPO algorithm under the TRL framework, with LoRA configured as rank 16, $\alpha=32$, and dropout rate 0.05. Due to resource constraints, we used a random subset of 1,000 training samples for the training. All evaluations across both settings are performed in a zero-shot manner.

For both direct training and \textit{re-cognizing optimization}, we use the Paged AdamW optimizer with a learning rate of 2e-4, $\beta_1$ = 0.9, $\beta_2$ = 0.95, perform gradient clipping at 0.3 with a warmup ratio of 10\%, set the global batch size to 16 (8 per device $\times$ 2 accumulation steps), and conduct bfloat16 mixed - precision training.

For GRPO experiments, we use the AdamW optimizer with a learning rate of 5e-5, perform gradient clipping at 0.3, apply a 15\% warmup over 100 steps, set weight decay to 0.01, use a global batch size of 8 (2 per device × 4 accumulation steps), enable fp16 mixed-precision training, employ a cosine-with-restarts learning-rate schedule, generate 4 completions per prompt with a maximum length of 512 tokens at temperature 0.9, and scale rewards.

\paragraph{Compute Resources}
Our \textit{re-cognizing optimization} experiments were run on eight NVIDIA RTX 4090d GPUs (though a single 4090d can support smaller runs), with each full training run taking approximately 6 hours. GRPO-based RL experiments used four NVIDIA A800 GPUs (minimum one A800 required) and averaged around 10 hours per run.

\subsection{Artifact License and Terms of Use.}
Our work leverages existing public logical reasoning benchmarks (Reclor, LogiQA, LogiQA2.0, LogicBench) and open-source models (LLaMA3.1-8B, Qwen2.5-7B) for experiments, adhering to their original academic use permissions. The anonymized code for SLKD calculation and re-cognizing optimization is released via a public repository (\url{https://anonymous.4open.science/r/Logical_Structure_as_Knowledge-81B1/}) for non-commercial research purposes only. We do not claim ownership of any original dataset or model content, and all derived artifacts (e.g., SLKD-scored samples) are restricted to research use in line with the source materials’ terms.

\subsection{Consistency with Intended Use.}
The existing datasets employed in this work are originally designed for evaluating LLM logical reasoning capabilities, which aligns with our research goal of enhancing model reasoning via structured logical knowledge. Our re-cognizing optimization strategy uses these datasets to validate the effectiveness of SLKD, without repurposing them for unintended scenarios. For the artifacts we create (SLKD metric, optimization framework), we specify their intended use as research tools for data-centric LLM reasoning enhancement, ensuring compatibility with the original access conditions of source datasets.

\subsection{Personally Identifying Information and Offensive Content.}
The benchmark datasets used in our experiments are standard, curated NLP resources widely adopted for logical reasoning research, which do not contain intentional personally identifying information or offensive content. Our logical element extraction and SLKD calculation processes focus solely on decomposing textual logical structures (predicates, constants, logical expressions) and do not introduce, amplify, or expose sensitive attributes. All data processing steps prioritize preserving the original dataset’s safety and compliance, with no use of personally identifiable or offensive content in experiments.

\subsection{Examples of Different Score Data}
\label{append:example}


\begin{tcolorbox}[
    breakable,title=Examples of Different Score Data,
    fontupper=\small,
]
\columnseprule=0.5pt
\sloppy

"context": "john knows how to play the piano"

"question": "does this entail that someone has the ability to play the piano?"

"answer": "yes"

"ReasoningIntensityScore": 0.05729995978985077

——————

"context": "Roves had held a senior position in the Navy before taking office.One of his good friends asked him about the Navy's plan to establish a submarine base on an island.Roosevelt looked around mysteriously and asked in a low voice.\"Can you keep it a secret?" "Of course I can!" The friend was very sure."So," Roosevelt said with a smile, "I can too.""

"question": "This text tells us:"

"options": ["Detours can also achieve the goal."

"Humor can subtly solve problems"

"Adherence to principles and flexibility are not contradictory."

"Don't do anything to others"],

"answer": "1"

"ReasoningIntensityScore": 0.589367067922439

——————

"context": "It is necessary to pay attention to avoiding hollowing out in the development of the service industry, but it is wrong and dangerous to think that the rapid development of modern service industry in China's economic growth will definitely lead to a hollowing out of the industry.This view of China will make China's economy lose an important window period for the rapid development of the modern service industry.In fact, the formation of an industrial structure dominated by the service industry does not mean the decline of the status of the manufacturing industry, nor does it mean "de-industrialization" "It is not the same as starting the hollowing out process of the industry."

"question": "The main emphasis of this text?"

"options": ["The rapid development of modern service industry cannot lead to a hollowing out of the industry"

"How to objectively evaluate the advantages and disadvantages of the rapid development of modern service industry"

"Whether it will cause industrial hollowing depends on the prosperity of the manufacturing industry"

"Don't worry about the hollowness of the industry and miss the opportunity to develop the service industry"],

"answer": "3"

"ReasoningIntensityScore": 0.9464210784935705

\end{tcolorbox}


\subsection{Reasoning Intensity Score Calculation Prompt}
\label{app:know_sys_con_prompt}
\begin{tcolorbox}[breakable,title=Prompt for Logical Decomposition]
\columnseprule=0.5pt


Instructions: Please extract the predicates and constants from the following context and create logical expressions that represent the relationships described. Format the output as a dictionary where each entry is a list of items. Follow these specific rules:

1. **Extract predicates** as core action words or relationships defining connections between entities, and output them as a list under the 'Predicates' key.

2. **Extract constants** as the specific entities or values mentioned in the context, and output them as a list under the 'Constants' key.

3. **Create logical expressions** using the extracted predicates and constants. Each logical expression should be simple and based on a single predicate. Output them as a list under the 'Logical Expressions' key.

4. Ensure that the output strictly follows the format provided in the example.

5. Do not combine expressions using logical operators such as 'and,' 'or,' etc., unless the relationship is explicitly mentioned in the context.


Example:

[Context: If an individual consumes a significant amount of water, they will experience a state of hydration. Conversely, if excessive amounts of sugar are ingested, a sugar crash will ensue. It is known that at least one of the following statements is true: either Jane consumes ample water or she will not experience a sugar crash. However, the actual veracity of either statement remains ambiguous, as it could be the case that only the first statement is true, only the second statement is true, or both statements are true.]

[Output:

Predicates: [`Consumes(x, y)`: Represents the act of 'x' consuming 'y' (e.g., an individual consuming water or sugar).,

`ExperienceState(x, y)`: Represents 'x' experiencing a state 'y' (e.g., hydration).,

`Ingested(x, y)`: Represents 'x' ingesting 'y' (e.g., excessive sugar).,

`Ensue(x)`: Represents that a condition 'x' follows or results (e.g., a sugar crash).,

`TrueStatement(x)`: Indicates that 'x' is known to be true.,

`NotExperience(x, y)`: Represents 'x' not experiencing a condition 'y' (e.g., not experiencing a sugar crash).],

Constants: [`Individual`: Represents a generic person in the context.,

`Water`: The substance being consumed by an individual.,

`Hydration`: The state that results from sufficient water consumption.,

`Sugar`: A substance that can be ingested.,

`SugarCrash`: The state that follows excessive sugar intake.,

`Jane`: A specific person mentioned in the context.],

Logical Expressions: [Consumes(Individual, Water)`,

`ExperienceState(Individual, Hydration)`,

`Ingested(Individual, Sugar)`,

`Ensue(SugarCrash)`,

`Consumes(Jane, Water)`,

`NotExperience(Jane, SugarCrash)`,

`\(\text{TrueStatement}(\text{Consumes}(\text{Jane}, \text{Water}) \lor \neg \text{Experience}(\text{Jane}, \text{SugarCrash}))\)`

---

** Tips for Extracting Predicates, Constants, and Logical Expressions:

- Focus on identifying core action or relational words for predicates.

- Extract constants as the specific entities mentioned.

- Use variables (`x`, `y`, etc.) to generalize when needed.

- Ensure logical expressions are complete and accurately reflect relationships.

** Your Task:

Context: [context]

\end{tcolorbox}

\begin{tcolorbox}[breakable,title=Prompt for Combinatorial Reasoning in BQA]
\columnseprule=0.5pt

Instructions: Please analyze the following binary question data. Since BQA is treated as an MCQA with a single option derived from the question, perform the analysis for this single option. Extract the relevant preconditions, define the deduction target, outline the deduction steps based on the provided Predicates, Constants, and Logical Expressions, and determine whether the option is correct based on the given answer index. Follow these specific rules:

1. **Output** should be a JSON object containing only the `option\_analysis` field, which is a list of analyses for each option.

2. For the option\_analysis, include the following fields:

   - option\_index: The index of the option (0-based).
   
   - option\_text: The text of the option (extracted from the question).
   
   - preconditions: A list of relevant preconditions from the Logical Expressions that pertain to the option.
   
   - deduction\_target: The abstracted logical conclusion that the option is attempting to establish.
   
   - deduction\_steps: A step-by-step logical deduction process from the preconditions to the deduction target. Each step should include:
   
     - step: The step number.
     
     - task: A description of what is being checked or inferred in this step.
     
     - expression: The logical expression used in this step, enclosed in backticks.
     
     - result: The outcome of this step, enclosed in backticks.
     
     - If a deduction step cannot proceed due to unsupported premises, indicate the failure and terminate further steps.
     
   - is\_correct: A boolean indicating whether the option is correct (true) or not (false). This should align with the answer field. When 'answer' is equal to 'yes', the value is true, otherwise the value is false.
   
3. **Format Requirements**:

   - The output must strictly follow the JSON structure as shown in the example.
   - Ensure consistency in field naming and hierarchy.
   
   - Do not include any additional fields or information not specified in the example.
   
4. **Important Considerations**:

   - Only include preconditions that are directly relevant to the option being analyzed.
   
   - Maintain logical rigor in deduction steps, ensuring each step follows from the previous ones based on the preconditions.
   - Avoid including unrelated preconditions to minimize complexity and enhance clarity.

Example:

- context: All people who regularly drink coffee are dependent on caffeine. People either regularly drink coffee or joke about being addicted to caffeine. No one who jokes about being addicted to caffeine is unaware that caffeine is a drug. Rina is either a student and unaware that caffeine is a drug, or neither a student nor unaware that caffeine is a drug. If Rina is not a person dependent on caffeine and a student, then Rina is either a person dependent on caffeine and a student, or neither a person dependent on caffeine nor a student.

- question: Rina is a person who jokes about being addicted to caffeine or unaware that caffeine is a drug.

- options: Rina is a person who jokes about being addicted to caffeine or unaware that caffeine is a drug.
  
- answer: yes

- Predicates:

  - RegularlyDrink(x, y): Represents 'x' regularly drinking 'y' (e.g., a person regularly drinking coffee).
  
  - DependentOn(x, y): Represents 'x' being dependent on 'y' (e.g., a person dependent on caffeine).
  
  - JokeAbout(x, y): Represents 'x' joking about 'y' (e.g., a person joking about being addicted to caffeine).
  
  - UnawareThat(x, y): Represents 'x' being unaware that 'y' (e.g., a person unaware that caffeine is a drug).
  
  - IsStudent(x): Represents 'x' being a student.
  
  - IsNeither(x, y): Represents 'x' being neither 'y' (used for expressing negation of multiple conditions).
  
- Constants:

  - People: Generic individuals.
  
  - Coffee: The beverage being consumed.
  
  - Caffeine: The substance people can be dependent on.
  
  - Rina: A specific person mentioned in the context.
  
- Logical Expressions:

  - \( \text{DependentOn}(\text{People}, \text{Caffeine}) \Rightarrow \text{RegularlyDrink}(\text{People}, \text{Coffee}) \)
  
  - \( \text{JokeAbout}(\text{People}, \text{Caffeine}) \vee \text{RegularlyDrink}(\text{People}, \text{Coffee}) \)
  
  - \( \text{UnawareThat}(\text{People}, \text{Caffeine}) \Rightarrow \text{JokeAbout}(\text{People}, \text{Caffeine}) \)
  
  - \( \text{IsStudent}(\text{Rina}) \wedge \text{UnawareThat}(\text{Rina}, \text{Caffeine}) \vee \neg \text{IsStudent}(\text{Rina}) \wedge \neg \text{UnawareThat}(\text{Rina}, \text{Caffeine}) \)
  
  - \( \neg \text{DependentOn}(\text{Rina}, \text{Caffeine}) \wedge \text{IsStudent}(\text{Rina}) \Rightarrow (\text{DependentOn}(\text{Rina}, \text{Caffeine}) \wedge \text{IsStudent}(\text{Rina})) \vee \neg (\text{DependentOn}(\text{Rina}, \text{Caffeine}) \wedge \text{IsStudent}(\text{Rina})) \)

Output:

option\_analysis: 

- **Option Index**: 0

- **Option Text**: Rina is a person who jokes about being addicted to caffeine or unaware that caffeine is a drug.

- **Preconditions**:

  - \( \text{JokeAbout}(\text{People}, \text{Caffeine}) \vee \text{RegularlyDrink}(\text{People}, \text{Coffee}) \)
  
  - \( \text{UnawareThat}(\text{People}, \text{Caffeine}) \Rightarrow \text{JokeAbout}(\text{People}, \text{Caffeine}) \)
  
- **Deduction Target**: \( \text{JokeAbout}(\text{Rina}, \text{Caffeine}) \vee \text{UnawareThat}(\text{Rina}, \text{Caffeine}) \)

- **Deduction Steps**:

  1. **Step**: 1
  
     - **Task**: Instantiate the general disjunction for Rina from the population-level statement.
     
     - **Expression**: \( \text{JokeAbout}(\text{Rina}, \text{Caffeine}) \vee \text{RegularlyDrink}(\text{Rina}, \text{Coffee}) \)
     
     - **Result**: Derived from \( \text{JokeAbout}(\text{People}, \text{Caffeine}) \vee \text{RegularlyDrink}(\text{People}, \text{Coffee}) \)
     
  2. **Step**: 2
  
     - **Task**: Apply the implication that joking about caffeine addiction leads to being unaware that caffeine is a drug for Rina.
     
     - **Expression**: \( \text{UnawareThat}(\text{Rina}, \text{Caffeine}) \Rightarrow \text{JokeAbout}(\text{Rina}, \text{Caffeine}) \)
     
     - **Result**: If Rina jokes about caffeine, then Rina is unaware that caffeine is a drug.
     
  3. **Step**: 3
  
     - **Task**: Combine the instantiated disjunction with the implication to derive the final conclusion.
     
     - **Expression**: \( \text{JokeAbout}(\text{Rina}, \text{Caffeine}) \vee \text{UnawareThat}(\text{Rina}, \text{Caffeine}) \)
     
     - **Result**: Since \( \text{JokeAbout}(\text{Rina}, \text{Caffeine}) \) implies \( \text{UnawareThat}(\text{Rina}, \text{Caffeine}) \), the disjunction holds. 

- **Is Correct**: True 

Tips for Option Analysis:

- **Preconditions**: Only include logical expressions that are directly relevant to the option being analyzed. Avoid listing all possible preconditions.

- **Deduction Steps**: Ensure each step logically follows from the previous one based on the preconditions. If a step cannot be completed due to insufficient support from the preconditions, indicate the failure and stop further deductions for that option.

- **is\_correct**: This field should be true only for the option that matches the answer field. Since BQA has only one option, is\_correct should align with the answer field.
- **Format Consistency**: Maintain the same JSON structure and field naming conventions across all options to ensure uniformity and ease of data extraction.

- **Logical Accuracy**: Ensure that all logical expressions and deductions accurately reflect the relationships defined by the predicates and constants.

Your Task:

Analyze the following Input data and generate the option\_analysis section as per the example above. Replace the xxx placeholders in the example with actual data derived from the input.

Input Data:

input\_data\_here

Please generate the option\_analysis section based on the above input data.

\end{tcolorbox}

\begin{tcolorbox}[breakable,title=Prompt for Combinatorial Reasoning in MCQA]
\columnseprule=0.5pt

Instructions:
Please analyze the following multiple-choice question data. For each option, extract the relevant preconditions, define the deduction target, outline the deduction steps based on the provided Predicates, Constants, and Logical Expressions, and determine whether the option is correct according to the given answer index. Follow these specific rules:

1. The **Output** should be a JSON object containing only the `option\_analysis` field, which is a list of analyses for each option.

2. For each option in `option\_analysis`, include the following fields:

   - **option\_index**: The index of the option (starting from 0).
   
   - **option\_text**: The text content of the option.
   
   - **preconditions**: A list of relevant preconditions from the Logical Expressions that pertain to the option.
   
   - **deduction\_target**: The abstracted logical conclusion that the option is attempting to establish.
   
   - **deduction\_steps**: A step-by-step logical deduction process from the preconditions to the deduction target. Each step should include:
   
     - **step**: The step number.
     
     - **task**: A description of what is being checked or inferred in this step.
     
     - **expression**: The logical expression used in this step, enclosed in backticks. Here, logical symbols like "implies" is represented as "\(\Rightarrow\)", "and" as "\(\land\)", "or" as "\(\lor\)", "not" as "\(\neg\)", "for all" as "\(\forall\)", "there exists" as "\(\exists\)" in LaTeX notation.
     
     - **result**: The outcome of this step, enclosed in backticks.
     
     - If a deduction step cannot proceed due to unsupported premises, indicate the failure and terminate further steps for that option.
     
   - **is\_correct**: A boolean indicating whether the option is correct (true) or not (false). The values of 'Answer' are '0', '1', '2', '3', where '0' represents the first option, '1' represents the second option, and so on. If the value of 'option\_index' is the same as the value of 'Answer', then the value of 'is\_correct' is 'true'.
   
3. **Format Requirements**:

   - The output must strictly follow the JSON structure as shown in the example.
   - Ensure consistency in field naming and hierarchy.
   
   - Do not include any additional fields or information not specified in the example.
   
4. **Important Considerations**:

   - Only include preconditions that are directly relevant to the option being analyzed.
   
   - Maintain logical rigor in deduction steps, ensuring each step follows from the previous ones based on the preconditions.
   
   - Avoid including unrelated preconditions to minimize complexity and enhance clarity.

Example:

- **Context**: In rheumatoid arthritis, the body's immune system misfunctions by attacking healthy cells in the joints causing the release of a hormone that in turn causes pain and swelling. This hormone is normally activated only in reaction to injury or infection. A new arthritis medication will contain a protein that inhibits the functioning of the hormone that causes pain and swelling in the joints.
- **Question**: The statements above, if true, most strongly support which one of the following conclusions?

- **Options**:
  1. Unlike aspirin and other medications that reduce pain and swelling and that are currently available, the new medication would repair existing cell damage that had been caused by rheumatoid arthritis.
  
  2. A patient treated with the new medication for rheumatoid arthritis could sustain a joint injury without becoming aware of it.
  
  3. Joint diseases other than rheumatoid arthritis would not be affected by the new medication.
  
  4. The benefits to rheumatoid arthritis sufferers of the new medication would outweigh the medication's possible harmful side effects.
  
- **Answer**: 1

- **Predicates**:
  - \( \text{Attack}(x, y) \): Represents 'x' attacking 'y' (e.g., the immune system attacking healthy cells).
  
  - \( \text{Release}(x, y) \): Represents 'x' releasing 'y' (e.g., the release of a hormone).
  
  - \( \text{Cause}(x, y) \): Represents 'x' causing 'y' (e.g., the hormone causing pain and swelling).
  
  - \( \text{Activate}(x, y) \): Represents 'x' activating 'y' (e.g., the hormone being activated by injury or infection).
  
  - \( \text{Inhibit}(x, y) \): Represents 'x' inhibiting 'y' (e.g., the protein inhibiting the hormone).
  
  - \( \text{Contain}(x, y) \): Represents 'x' containing 'y' (e.g., the medication containing a protein).
  
- **Constants**:
  - \( \text{ImmuneSystem} \): The body's defense mechanism.
  
  - \( \text{HealthyCells} \): Cells in the joints that are not diseased.
  
  - \( \text{Hormone} \): A chemical messenger involved in causing pain and swelling.
  
  - \( \text{Pain} \): A sensation caused by the hormone.
  
  - \( \text{Swelling} \): A condition caused by the hormone.
  
  - \( \text{Injury} \): A condition that normally activates the hormone.
  
  - \( \text{Infection} \): A condition that normally activates the hormone.
  
  - \( \text{ArthritisMedication} \): A new medication for treating arthritis.
  
  - \( \text{Protein} \): A component of the medication that inhibits the hormone.
  
- **Logical Expressions**:
  - \( \text{Attack}(\text{ImmuneSystem}, \text{HealthyCells}) \)
  
  - \( \text{Release}(\text{ImmuneSystem}, \text{Hormone}) \)
  
  - \( \text{Cause}(\text{Hormone}, \text{Pain}) \)
  
  - \( \text{Cause}(\text{Hormone}, \text{Swelling}) \)
  
  - \( \text{Activate}(\text{Injury}, \text{Hormone}) \)
  
  - \( \text{Activate}(\text{Infection}, \text{Hormone}) \)
  
  - \( \text{Inhibit}(\text{Protein}, \text{Hormone}) \)
  
  - \( \text{Contain}(\text{ArthritisMedication}, \text{Protein}) \)

Option Analysis

1. **Option Index**: 0

   - **Option Text**: Unlike aspirin and other medications that reduce pain and swelling and that are currently available, the new medication would repair existing cell damage that had been caused by rheumatoid arthritis.
   
   - **Preconditions**:
   
     - \( \text{Attack}(\text{ImmuneSystem}, \text{HealthyCells}) \)
     
     - \( \text{Release}(\text{ImmuneSystem}, \text{Hormone}) \)
     
     - \( \text{Cause}(\text{Hormone}, \text{Pain}) \)
     
     - \( \text{Cause}(\text{Hormone}, \text{Swelling}) \)
   
   - **Deduction Target**: \( \text{Repair}(\text{ArthritisMedication}, \allowbreak \text{HealthyCellsDamage}) \)
   - **Deduction Steps**:
   
     - **Step 1**:
     
       - **Task**: Check if \( \text{Attack}(\text{ImmuneSystem}, \text{HealthyCells}) \) implies \( \text{Damage}(\text{HealthyCells}) \).
       
       - **Expression**: \( \text{Attack}(\text{ImmuneSystem}, \text{HealthyCells}) \Rightarrow \text{Damage}(\text{HealthyCells}) \)
       
       - **Result**: Supported by context (immune system attacking healthy cells causes damage).
       
     - **Step 2**:
     
       - **Task**: Check if \( \text{Contain}(\text{ArthritisMedication}, \text{Protein}) \) and \( \text{Inhibit}(\text{Protein}, \text{Hormone}) \) imply \( \text{Repair}(\text{ArthritisMedication}, \allowbreak \text{HealthyCellsDamage}) \).

       - **Expression**: \( \text{Contain}(\text{ArthritisMedication}, \text{Protein}) \land \text{Inhibit}(\text{Protein}, \text{Hormone}) \Rightarrow \text{Repair}(\text{ArthritisMedication}, \allowbreak \text{HealthyCellsDamage}) \)
       
       - **Result**: Not supported. The context only states that the protein inhibits the hormone, not that it repairs damage.
     
     - **Step 3**:
      
       - **Task**: Derivation fails.
     
       - **Expression**: Derivation cannot 
       proceed.
     
       - **Result**: \( \text{Repair}(\text{ArthritisMedication}, \allowbreak \text{HealthyCellsDamage}) \) cannot be derived from the given preconditions.
   
   - **Is Correct**: False

2. **Option Index**: 1

   - **Option Text**: A patient treated with the new medication for rheumatoid arthritis could sustain a joint injury without becoming aware of it.

   - **Preconditions**:

     - \( \text{Cause}(\text{Hormone}, \text{Pain}) \)

     - \( \text{Activate}(\text{Injury}, \text{Hormone}) \)

     - \( \text{Inhibit}(\text{Protein}, \text{Hormone}) \)

     - \( \text{Contain}(\text{ArthritisMedication}, \text{Protein}) \)

   - **Deduction Target**: \( \exists \text{Patient}, \text{Injury} : [\text{Sustain}(\text{Patient}, \text{Injury}) \land \text{Unaware}(\text{Patient}, \text{Injury})] \)

   - **Deduction Steps**:

     - **Step 1**:
 
       - **Task**: Determine the effect of \( \text{Inhibit}(\text{Protein}, \text{Hormone}) \) from the medication.
 
       - **Expression**: \( \text{Inhibit}(\text{Protein}, \text{Hormone}) \)
 
       - **Result**: Supported by context: The protein inhibits the hormone that causes pain and swelling.
 
     - **Step 2**:

       - **Task**: Analyze the implication of inhibiting the hormone on pain and swelling.
  
       - **Expression**: \( \text{Inhibit}(\text{Protein}, \text{Hormone}) \Rightarrow \neg \text{Cause}(\text{Hormone}, \text{Pain}) \land \neg \text{Cause}(\text{Hormone}, \text{Swelling}) \)
   
       - **Result**: Supported by context: If the hormone is inhibited, it cannot cause pain and swelling.
    
     - **Step 3**:
    
       - **Task**: Infer the patient's awareness of injury when pain and swelling are absent.
     
       - **Expression**: \( \neg \text{Cause}(\text{Hormone}, \text{Pain}) \land \neg \text{Cause}(\text{Hormone}, \text{Swelling}) \Rightarrow \text{Unaware}(\text{Patient}, \text{Injury}) \)
    
       - **Result**: Supported by context: Without pain and swelling, the patient may not be aware of sustaining an injury.
    
     - **Step 4**:
    
       - **Task**: Combine the above implications to conclude the deduction target.
     
       - **Expression**: \( \exists \text{Patient}, \text{Injury} : [\text{Sustain}(\text{Patient}, \text{Injury}) \land \text{Unaware}(\text{Patient}, \text{Injury})] \)
       - **Result**: Deduction is valid based on the inhibited hormone preventing awareness of injury.
  
   - **Is Correct**: True

3. **Option Index**: 2

   - **Option Text**: Joint diseases other than rheumatoid arthritis would not be affected by the new medication.
  
   - **Preconditions**:
  
     - \( \text{Inhibit}(\text{Protein}, \text{Hormone}) \)
 
     - \( \text{Contain}(\text{ArthritisMedication}, \text{Protein}) \)

   - **Deduction Target**: \( \forall x [\text{JointDisease}(x) \land x \neq \text{RheumatoidArthritis} \Rightarrow \neg \text{Affect}(\text{ArthritisMedication}, x)] \)
 
   - **Deduction Steps**:

     - **Step 1**:

       - **Task**: Check if the context provides information about other joint diseases.
 
       - **Expression**: \( \text{JointDisease}(x) \land x \neq \text{RheumatoidArthritis} \)
 
       - **Result**: Not supported. The context only discusses rheumatoid arthritis.
 
     - **Step 2**:

       - **Task**: Determine if there is any implication that the medication specifically targets rheumatoid arthritis.
 
       - **Expression**: \( \text{Contain}(\text{ArthritisMedication}, \text{Protein}) \Rightarrow \text{SpecificEffect}(\text{RheumatoidArthritis}) \)
 
       - **Result**: Not supported. The context does not specify that the protein exclusively affects rheumatoid arthritis.

     - **Step 3**:

       - **Task**: Derivation fails.
 
       - **Expression**: Derivation cannot proceed.
 
       - **Result**: Cannot conclude that the medication does not affect other joint diseases.

   - **Is Correct**: False

4. **Option Index**: 3

   - **Option Text**: The benefits to rheumatoid arthritis sufferers of the new medication would outweigh the medication's possible harmful side effects.

   - **Preconditions**:

     - \( \text{Cause}(\text{Hormone}, \text{Pain}) \)

     - \( \text{Cause}(\text{Hormone}, \text{Swelling}) \)

     - \( \text{Inhibit}(\text{Protein}, \text{Hormone}) \)

     - \( \text{Contain}(\text{ArthritisMedication}, \text{Protein}) \)

   - **Deduction Target**: \( \text{Benefit}(\text{ArthritisMedication}) > \text{HarmfulSideEffect}(\text{ArthritisMedication}) \)

   - **Deduction Steps**:

     - **Step 1**:

       - **Task**: Identify the benefits of the medication based on inhibiting the hormone.

       - **Expression**: \( \text{Inhibit}(\text{Protein}, \text{Hormone}) \Rightarrow \text{Reduce}(\text{Pain}) \land \text{Reduce}(\text{Swelling}) \)

       - **Result**: Supported by context: The protein inhibits the hormone, which causes pain and swelling.

     - **Step 2**:

       - **Task**: Determine if the context provides information about harmful side effects.

       - **Expression**: \( \text{HarmfulSideEffect}(\text{ArthritisMedication}) \)

       - **Result**: Not supported. The context does not mention any side effects of the medication.

     - **Step 3**:

       - **Task**: Derivation fails.

       - **Expression**: Derivation cannot proceed.

       - **Result**: Cannot compare benefits and harmful side effects due to lack of information on side effects.

   - **Is Correct**: False

Tips for Option Analysis

- **Preconditions**: Only include logical expressions that are directly relevant to the option being analyzed. Avoid listing all possible preconditions.

- **Deduction Steps**: Ensure each step logically follows from the previous one based on the preconditions. If a step cannot be completed due to insufficient support from the preconditions, indicate the failure and stop further deductions for that option.

- **is\_correct**: This field should be true only for the option that matches the answer index. All other options should be false.

- **Format Consistency**: Maintain the same JSON structure and field naming conventions across all options to ensure uniformity and ease of data extraction.

- **Logical Accuracy**: Ensure that all logical expressions and deductions accurately reflect the relationships defined by the predicates and constants.

Your Task

Analyze the following Input data and generate the option\_analysis section as per the example above. Replace the xxx placeholders in the example with actual data derived from the input.

Input Data

input\_data\_here

Please generate the option\_analysis section based on the above input data. 

\end{tcolorbox}

\begin{tcolorbox}[breakable,title=Prompt for GRPO training in BQA]
\columnseprule=0.5pt

You are solving a yes/no (BQA) problem. Follow these rules exactly:
1. First, think through your reasoning and enclose it in <think> and </think>.
2. Then, give your final answer ("yes" or "no") enclosed in <answer> and </answer>.
3. Do not output anything else; any deviation yields zero reward.
4. Your <think> block should not exceed 450 tokens.
5. Do not include <answer> inside your <think>, and vice versa.
6. Structure your reasoning in three steps in <think> block:
Step 1: Analyze the Context.
Step 2: Analyze the Question.
Step 3: Deduce the Conclusion.

Example:
Context: If Noah takes a vacation, then he will relax. If Noah works overtime, then he will not have enough time to relax. We know at least one of the following is true: he won't relax or he will have enough time to relax. We do not know which.
Question: If at least one of these statements is true, can we conclude that essentially "Noah takes a vacation" or "Noah works overtime" must be true?

Your response:
<think>Step 1: Context Analysis: Let V="takes vacation", O="works overtime", R="relaxes", T="has enough time to relax". The context gives implications "if V is true, then R is true" and "if O is true, then T is false", and the disjunction "R is false" or "T is true".
Step 2: Question Analysis: The question asks whether V or O must always be true given ?R or T.
Step 3: Answer analysis: First, from "if V is true, then R is true", we can infer that "if R is false, then V is false". From "if O is true, then T is false", we can infer that "if T is true, then O is false". So, "R is false" or "T is true" implies that "V is false" or "O is false". Thus, V and O cannot both be true, and it's possible for both V and O to be false. Therefore, it's not certain that V or O must be true.</think><answer>no</answer>

Now solve this BQA problem:

Context: 

Question:

Your response:

\end{tcolorbox}

\begin{tcolorbox}[breakable,title=Prompt for GRPO training in MCQA]
\columnseprule=0.5pt
You are solving a multiple-choice (MCQA) problem with four options (A-D). Follow these rules exactly:
1. First, think through your reasoning and enclose it in <think> and </think>.
2. Then, give your final answer ("A", "B", "C", or "D") enclosed in <answer> and </answer>.
3. Do not output anything else; any deviation yields zero reward.
4. Your <think> block should not exceed 450 tokens.
5. Do not include <answer> inside your <think>, and vice versa.
6. Structure your reasoning in three steps in <think> block:
Step 1: Analyze the Context.
Step 2: Analyze the Question.
Step 3: Analyze the Options and deduce the best choice.

Example:
Context: In recent years, many cabinetmakers have been winning acclaim as artists. But furniture must be useful, so cabinetmakers focus on utility, implying cabinetmaking is not art.
Question: Which assumption supports the conclusion that cabinetmaking is not art?
Options:
A. Some furniture is made purely for display.
B. Artists are not concerned with monetary value.
C. Cabinetmakers should focus more on practical utility.
D. Paying attention to utility disqualifies an object as art.

Your response:
<think>Step 1: Context Analysis: The passage states that because furniture must be useful, cabinetmakers must prioritize utility, so their work cannot be art.
Step 2: Question Analysis: We need the hidden premise that links utility focus to art classification.
Step 3: Options Analysis: Option A: Irrelevant; museums display does not address utility vs art.
Option B: Off-topic; monetary concern is not mentioned.
Option C: Restates the problem but does not link utility to disqualification of art.
Option D: Directly asserts that focusing on utility means an object is not art, exactly matching the conclusion.
Choice: Option D is the clear support.</think><answer>D</answer>

Now solve this MCQA problem:

Context: 

Question: 

Options:

Your response:

\end{tcolorbox}

\end{document}